\documentclass[11pt]{article}
\usepackage{coling2020}
\usepackage{times}
\usepackage{url}
\usepackage{float}
\usepackage{latexsym}
\usepackage{graphicx}
\usepackage{caption}
\usepackage{hyperref}
\usepackage{svg}
\usepackage{amsmath, amssymb, amsfonts, multirow, bm}
\usepackage{pifont}
\newcommand{\cmark}{\ding{51}}%
\newcommand{\xmark}{\ding{55}}%
\usepackage{subcaption}

\colingfinalcopy 

\title{Event-Guided Denoising for Multilingual Relation Learning}

\author{Amith Ananthram \And Emily Allaway \\
  Department of Computer Science, Columbia University, New York, NY \\
  {\tt amith.ananthram@columbia.edu, \tt \{eallaway,kathy\}@cs.columbia.edu}
  \And Kathleen McKeown 
 }

\date{}

\begin{document}
\maketitle
\begin{abstract}
General purpose relation extraction has recently seen considerable gains in part due to a massively data-intensive distant supervision technique from~\newcite{Soares:19} that produces state-of-the-art results across many benchmarks. In this work, we present a methodology for collecting high quality training data for relation extraction from unlabeled text that achieves a near-recreation of their zero-shot and few-shot results at a fraction of the training cost. Our approach exploits the predictable distributional structure of date-marked news articles to build a denoised corpus -- the extraction process filters out low quality examples. We show that a smaller multilingual encoder trained on this corpus performs comparably to the current state-of-the-art (when both receive little to no fine-tuning) on few-shot and standard relation benchmarks in English and Spanish despite using many fewer examples ($50k$ vs. $300$mil+).
\end{abstract}

\section{Introduction}
\label{intro}

%
%
\blfootnote{
    %
    %
    %
    %
    %
    \hspace{-0.65cm}  
    This work is licensed under a Creative Commons 
    Attribution 4.0 International License.
    License details:
    \url{http://creativecommons.org/licenses/by/4.0/}.
}

\begin{figure}
\centering
  \includegraphics[width=\linewidth]{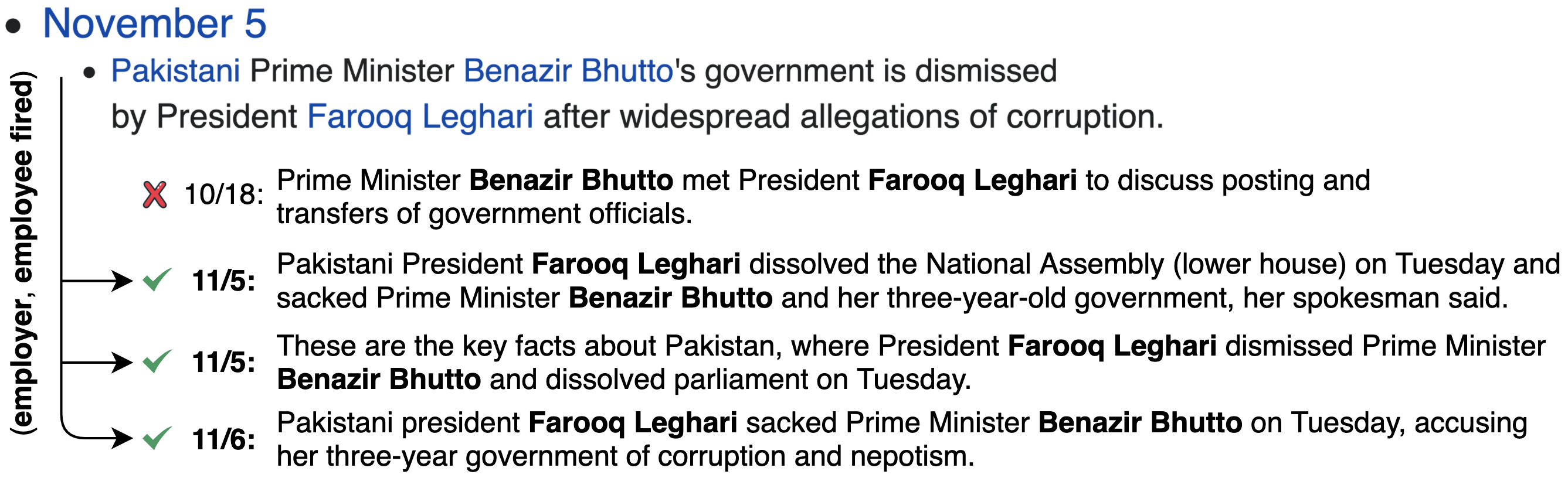}
  \caption{\textit{event-guided} denoising: using date-marked events from Wikipedia to select a diverse group of sentences that describe the same relation (\textcolor{green}{\cmark}) from news articles, filtering out noisy examples (\textcolor{red}{\xmark})}
  \label{fig:denoising}
\end{figure}

Multilingual relation extraction is an important problem in NLP, facilitating a diverse set of downstream tasks from the autopopulation of knowledge graphs (e.g.,~\newcite{Ji:2011}) to question answering (e.g.,~\newcite{Xu:2016}). While early efforts in relation extraction used supervised methods that rely on a fixed set of predetermined relations, research has since shifted to the identification of arbitrary unseen relations in any language. In this paper, we present a method for extracting high quality relation training examples from date-marked news articles. This technique leverages the predictable distributional structure of such articles to build a corpus that is denoised (i.e., where sentences with the \textit{same} entities only express the \textit{same} relation, see Figure~\ref{fig:denoising}). We use this corpus to learn general purpose relation representations and evaluate their quality on few-shot and standard relation extraction benchmarks in English and Spanish with little to no task-specific fine-tuning, achieving comparable results to a significantly more data-intensive approach that is the current state-of-the-art.

The current state-of-the-art model, ``Matching the Blanks" or MTB~\cite{Soares:19}, is a distant supervision~\cite{Mintz:2009} technique that provides large gains on many relation extraction benchmarks and builds on Harris' distributional hypothesis~\cite{Harris:54} and its extensions~\cite{Lin:2001}. ~\newcite{Soares:19} assume that the informational redundancy of very large text corpora (e.g., Wikipedia) results in sentences that contain the same pair of entities generally expressing the same relation. Thus, an encoder trained to collocate such sentences can be used to identify the relation between entities in any sentence $s$ by finding the labeled relation example whose embedding is closest to $s$. While~\newcite{Soares:19} achieve state-of-the-art on FewRel~\cite{Han:2018} and SemEval 2010 Task 8~\cite{Hendrickx:2019}, their approach relies on a huge amount of data, making it difficult to retrain in English or any other language with standard computational resources: they fine-tune BERT large~\cite{Devlin:2019}, which has $340$mil parameters, on $300$mil+ relation pair statements with a batch size of $2,048$ for $1$mil steps. In contrast our method, with only $50k$ relations statements and a language-model one-third the size, achieves comparable performance when fine-tuned on little to no task-specific data. 

Our main contribution is a distant supervision approach in which we assume that sections of news corpora exhibit even more informational redundancy than Wikipedia. Specifically, news in the days following an event (e.g., the 2006 World Cup) frequently re-summarizes the event before adding new details. As a result, news exhibits a strong form of local consistency over short rolling time windows where otherwise fluid relations between entities remain fixed. For example, the relation between Italy and France as expressed in a random piece of text is dynamic and context-dependent, spanning a wide range of possibilities that include ``enemies", ``neighbors" and ``allies".  But, in the news coverage following the 2006 World Cup, it is static -- they are sporting competitors. Therefore, by considering only sentences around specific events, we extract groups of statements that express the same relation and are relatively free of noise (i.e. statements that express a different relation, see Figure~\ref{fig:denoising}).  

Training multilingual BERT (mBERT) on our denoised corpus yields relation representations that adapt well to resource-constrained downstream tasks: we evaluate their quality on FewRel and SemEval 2010 Task 8, producing near state-of-the-art results when finetuned on little to no task-specific data. In addition to the strong performance of our approach in English, it is easily generalizable to other languages, requiring only news corpora and event descriptions from Wikipedia to build a high-quality training corpus. We evaluate this in Spanish and find our method outperforms mBERT on the TAC KBP 2016 relation corpus~\cite{Ellis:2015}. We share our code to allow other researchers to apply our approach to news corpora of their own.\footnote{\url{https://github.com/amith-ananthram/event-guided-denoising}}

\section{Methods}
We focus on the task of learning to produce high quality general-purpose relation representations from relation statements in English and Spanish. We adopt the common definition of a relation statement in the literature: a triple $\bm{r} = (\bm{x}, \bm{s}_1, \bm{s}_2)$ where $\bm{x} = [x_0 \ ... \ x_n]$ is a sequence of tokens and $\bm{s}_1 = (i, j)$ and $\bm{s}_2 = (k, l)$ are the indices of special start and end identifier tokens that demarcate two entity mentions in $\bm{x}$. Our goal is to learn a function $f(\bm{r}) = h$ that maps a relation statement to a fixed-length vector $h \in \mathbb{R}^d$. The vector $h$ represents the relation between the entity mentions identified by $\bm{s}_1$ and $\bm{s}_2$ as expressed in $\bm{x}$. The cosine similarity between $f(\bm{r})$ and $f(\bm{r}_O)$ should be close to $1$ if and only if $\bm{r}$ and $\bm{r}_O$ express the same relation.  That is to say, $f$ should collocate sentences that exhibit similar relations.
\newline
\newline
\textbf{Data Denoising:} To learn such a function $f$, we build a corpus that contains groups of sentences that express similar relations. Following~\newcite{Soares:19}, we first extract all candidate relation statements from our news corpus by considering all pairwise combinations of entity mentions in every sentence. We then experiment with several different denoising techniques to select groups of similar relation statements from these candidates with the goal of filtering out dissimilar examples. These techniques can be decomposed into two separate steps: 1) pair selection, or how we select the entity pair for a given group, which includes \textit{event-guided} pairing, and 2) sentence selection, or how we select the sentences exhibiting the same relation for a selected entity pair.

\begin{table}
\begin{center}
\begin{tabular}{ | l | l | }
\hline
 \textbf{Step 1: Pair Selection Methods} & \textbf{Step 2: Sentence Selection Methods} \\ 
 \hline
 \textbf{\textit{Random}} (filtered by count and/or PPMI) & \textbf{\textit{Random}} \\
 \textbf{\textit{Date-window-grouped}}  (filtered by count and/or PPMI) & Only sentences within a \textbf{\textit{date window}} \\
 \textbf{\textit{Event-guided}} (filtered by count and/or PPMI) & \\
 \hline
\end{tabular}
\caption{Our two-step denoising approach.  To extract examples, we first select entity pairs with a pair selection method and then relation statements for those entity pairs with a sentence selection method.}
\label{tab:denoising}
\end{center}
\end{table}

\textit{Event-guided} pairing underpins our event-guided denoising technique. It is a three-step process that selects entity pairs related to a news event on a specific date. First, we collect a list of date-marked descriptions of events that occurred during the publication date range of our news corpus.  These descriptions are available in many languages for every calendar year and can be easily scraped from Wikipedia. Next, we extract all pairs of entities from these descriptions, grouping them by their event dates. Finally, we use these groupings to select pairs of entities from our corpus, filtering them by their count and in-article positive pointwise mutual information (PPMI,~\cite{Church:1990}) in the days following their event date. 

For comparison, we also evaluate \textit{random} and \textit{date-window-grouped} pairing. \textit{Random} pairing selects pairs of entities at random. \textit{Date-window-grouped} pairing groups pairs of entities by sliding date windows. As with \textit{event-guided} pairing, it is informed by our observation regarding the local consistency of relations between entities associated with events in news. However, it does not restrict the selected entities to those found in Wikipedia event descriptions. For both these methods, we filter pairs by their counts and PPMIs exactly as in \textit{event-guided} pairing.

Sentence selection is straightforward: for a selected entity pair, we select sentences containing the pair
either 1) at \textit{random} or 2) within a \textit{date window} (typically the dates used for the pair's selection).
\newline 
\newline 
\textbf{Training (Matching the Blanks):} We adopt the training methodology for learning relation representations employed by~\newcite{Soares:19} (MTB) in its entirety. That is, we train an encoder $f_\theta(\bm{r}) = h$ to produce relation representations $h \in R^d$ that determine whether two relation statements $\bm{r}_i$, $\bm{r}_j$ describe the same relation by computing their inner product.  As in~\newcite{Soares:19}, our relation representation $h$ is the concatenation of the final hidden state representations of the entity start tokens for the entity mentions identified by $\bm{s}_1$ and $\bm{s}_2$ in $\bm{r}$. 

For training, relation statements are collected into groups of positive examples (statements that align with the same relation type according to our denoising techniques) and negative examples (statements that align with a different relation type) for each selected entity pair. Following~\newcite{Soares:19}, negative examples include both ``easy" and ``hard" mentions.  ``Easy" mentions include no part of the entity pair and ``hard" mentions include exactly one of the entities in the entity pair, suggesting that these ``hard" examples describe similar but different relations that require disambiguation. With probability $\alpha$, each entity in a relation statement is replaced with a $[\text{BLANK}]$ token and with probability $\beta$ each token in a relation statement is replaced with a $[\text{MASK}]$ token. We train a parameterization of $f_\theta$ that minimizes both a masked language modeling loss~\cite{Taylor:1953,Devlin:2019} and a simple binary cross-entropy loss that encourages the representations of positive examples for a given entity pair to be closer to one another than to the negative examples.

\section{Experiments}

\begin{table}[h]
\centering
\begin{tabular}{ | c | r | r | r | r | r | }
\hline 
\textbf{Lang} & \textbf{\# Articles} & \textbf{\# Sentences} & \textbf{\# Entity Pairs} & \textbf{\# Rel. Statements} & \textbf{\# Denoised Statements} \\
\hline 
en & $810,000$ & $11,925,900$ & $32,831,624$ & $209,796,400$ & $189,281$\\
es & $98,534$ & $904,908$ & $9,315,975$ & $56,410,125$ & $95,881$ \\
\hline
\end{tabular}
\caption{RCV1 and RCV2 corpus summary statistics}
\label{tab:corpusstats}
\end{table}

\textbf{Data:} We work with the English RCV1 and the Spanish subset of the RCV2 Reuters news corpora~\cite{Lewis:04} due to their size and wide availability, extracting sentences from thousands of news articles from August 1996 to August 1997 (see Table \ref{tab:corpusstats}). We collect date-marked descriptions of events from Wikipedia for the same period.\footnote{For example, from \url{https://es.wikipedia.org/wiki/1996}} We extract nouns/named entities with spaCy\footnote{\url{https://spacy.io}} models. 

\noindent\newline
\textbf{Hyperparameters:} We fine-tune multilingual BERT, $f$,~\cite{Devlin:2019} on our English and Spanish corpora using Adam~\cite{Kingma:2015} with a learning rate of $3e-5$ and a batch size of $32$ on a single NVIDIA Tesla T4 GPU. We use a blank rate $\alpha = 0.7$, a mask rate $\beta = 0.15$, a BERT hidden layer dropout rate of $0.5$ and a sentence selection window of $+4$ days. We adopt the learning rate, blank rate and mask rate used by~\newcite{Soares:19}. For additional tuning details, see Appendix B.

\begin{table}[H]
\centering
\begin{tabular}{ | l | l | l | l | r | r | r | r |}
\hline
 \multirow{2}{*}{\textbf{Denoising}} & \multirow{2}{*}{\textbf{Model}} & 
 \multirow{2}{*}{\textbf{Langs}} & 
 \multirow{2}{*}{\textbf{\# Ex.}} & \multicolumn{2}{c|}{\textbf{$n=5, k=1$}} & \multicolumn{2}{c|}{\textbf{$n=10, k=1$}} \\ 
 \cline{5-8}
   &  &  & & en & es & en & es \\ 
 \hline
 & mBERT base & n/a & n/a & $62.9$ & $63.8$ & $58.7$ & $50.6$ \\
 & BERT base & n/a & n/a & $68.7$ & - & $57.1$ & - \\
 & BERT large & n/a & n/a & $72.9\dagger$ & - & $62.3\dagger$ & - \\
 & BERT large + MTB & en & $300$mil+ & $\bm{80.4}\dagger$ & - & $\bm{71.5}\dagger$ & - \\
 \hline
 \hline
date-window grouped & mBERT & en & $100k$ & $72.4$ & $67.5$ & $59.3$ & $56.7$ \\
 event-guided (EvtGD) & mBERT & en & $50$\textit{k} & $\bm{78.7}$ & $75.1$ & $\bm{69.0}$ & $64.3$ \\
  event-guided (EvtGD) & mBERT & en, es & $70$\textit{k} & $77.5$ & $\bm{76.7}$ & $68.1$ & $\bm{64.4}$ \\
 \hline
 
\end{tabular}
\caption{Zero-shot (no fine-tuning) accuracies on the FewRel development set. For each test case, $k$ is the number of relation types and $n$ is the number of examples per relation type. The development set is used for comparison with the $\dagger$'d results from MTB. Random pairing excluded due to poor performance.}
\vspace{-5mm}
\label{tab:fewrel}
\end{table}
\begin{table}[H]
    \begin{subtable}{0.55\linewidth}
    \centering
    \begin{tabular}{ | l | r | r | }
    \hline
     \textbf{Model} & \textbf{Corpus Size} & \textbf{1\%} \\ 
     \hline
     BERT large &  n/a & $28.6\dagger$ \\
     BERT large + MTB & $300$mil+ ex & $31.2\dagger$ \\
     \hline
     \hline
      \begin{tabular}[l]{@{}l@{}}mBERT\end{tabular} & n/a & $28.5$ \\
     \begin{tabular}[l]{@{}l@{}}mBERT + EvtGD\end{tabular} & $50k$ ex & $\bm{33.3}$ \\
     \hline
    \end{tabular}
    \caption{F1-scores on a random held out slice of 1500 examples from the SemEval 2010 Task 8 training set (for comparison with the $\dagger$'d results from MTB.).  Models are fine-tuned on 80 examples.}
    \label{tab:semeval}
    \end{subtable}\hfill
    \begin{subtable}{0.4\linewidth}
    \centering
    \begin{tabular}{| c | c | c | }
    \hline
     \textbf{Model} & \textbf{1\%} & \textbf{10\%} \\ 
     \hline
     mBERT & $33.0$ & $72.8$ \\
     \hline
     \hline
     \begin{tabular}[c]{@{}l@{}}mBERT\\+EvtGD\end{tabular} & $\bm{41.5}$ & $\bm{76.8}$ \\
     \hline
    \end{tabular}
    \caption{F1-scores on the TAC KBP 2016 Spanish relation corpus (from our en/es model, fine-tuned on 1\% and 10\% of its data and evaluated on the rest).}
    \label{tab:tackbp}
    \end{subtable}
    \caption{F1-scores on two standard supervised relation extraction benchmarks for English (SemEval 2010 Task 8) and Spanish (TAC KBP 2016). EvtGD indicates \textit{event-guided} denoising.}
\end{table}

\textbf{Evaluation:} We evaluate the performance of our embedding space on FewRel, SemEval 2010 Task 8 and on the Spanish subset of TAC KBP 2016 with little to no fine-tuning on task-specific data.

FewRel is a few-shot relation benchmark where, given a set of $n$ relation types with $k$ relation statement examples for each, the task is to identify which of the $n$ relation types is expressed by a query relation statement $\bm{q}$.  To do so, we embed $\bm{q}$ and all candidate relation statements $\bm{c}_{1}^{1}\hdots\bm{c}_{n}^k$, find the candidate $\bm{c}_{i}^{j}$ whose embedding is most similar to $\bm{q}$ and predict relation type $i$. In addition to English, we translate 10\% of FewRel's development set with Google Translate (which performs well on Spanish) \footnote{\url{translate.google.com}}.

SemEval 2010 Task 8 is a standard relation classification benchmark. TAC KBP 2016 is a Spanish knowledge base population corpus; though it is typically used to evaluate slot-filling tasks, it contains documents with labeled relations. For both corpora, following~\newcite{Soares:19} (including their hyperparameters which we include in Appendix C), we append individual dense layers to our embedding space to map relation representations $h \in \mathbb{R}^d$ to their respective classes. 

We compare our performance to the results published in~\newcite{Soares:19} (BERT large + MTB) and mBERT without \textit{event-guided} denoising.

\noindent\newline
\textbf{Results:} Our \textit{event-guided} denoising approach performs well with little to no fine-tuning on task-specific data. We achieve results comparable to or better than those of~\newcite{Soares:19} under similar fine-tuning constraints despite training a much smaller model ($110$mil vs $340$mil parameters) on much less training data ($50k$ vs $300$mil+  examples).  On FewRel, with no fine-tuning for either model, we achieve only a $2\%$ reduction in accuracy (on a benchmark where fine-tuning with even a small amount of data produces gains of over $5\%$) (Table \ref{tab:fewrel}).

On SemEval, when both models are fine-tuned on just 1\% (80 examples) of its training data, our model outperforms~\newcite{Soares:19} (Table \ref{tab:semeval}). These results align with an observation from their subsequent fine-tuning on the FewRel training set -- namely, what was most critical to improved performance was the number of different relation types seen during fine-tuning, not the number of examples for each relation type. Thus, we hypothesize that our event-guided approach is producing a small but diverse set of high-quality relations.

Our multilingual approach also outperforms mBERT on the FewRel (auto-translated) and TAC KBP Spanish subsets (Tables \ref{tab:fewrel}, \ref{tab:tackbp}). On TAC KBP, our approach achieves a higher F1 than mBERT when fine-tuned on the same small percentages of task-specific data.  After 20\% ($> 500$ examples), the two models' performances converge. 

These results, in English and Spanish, demonstrate the efficacy of our \textit{event-guided} denoising approach on resource-constrained relation extraction tasks.


\section{Conclusion}

We present an \textit{event-guided} denoising approach for relation extraction corpus creation that, when used with the current state-of-the-art training procedure, achieves comparable results in English under a low-resource regime for only a fraction of the training cost. It also performs well in Spanish, demonstrating its adaptability to resource-constrained relation extraction tasks in non-English languages. 

Our technique affords the broader research community the ability to approximate the current state-of-the-art in relation extraction by significantly lowering its associated training costs.  However, it requires a fairly large date-marked news corpus which may not be available in low resource languages. We leave an exploration of broader language coverage and minimal required corpus size for future work. 

One promising direction for expanding language coverage is cross-lingual learning via ``codeswitched" examples and other language modeling losses (e.g. \cite{Lample:2019}).  We hypothesize that such methods could help knowledge transfer among languages and improve results on downstream tasks.

Finally, we note that since our approach extracts relation statements from news corpora, it is likely that the resulting distribution of underlying relation types is different than the distribution found in Wikipedia. For example, Wikipedia may contain more expressions of standard ontological relations (e.g., \textit{father-of}, \textit{born-in}) characteristic of factoids. Despite this hypothesized difference, our approach performs well on both FewRel and SemEval 2010 Task 8, both of which include a subset of such relation types.  In the future we intend to investigate these differences and their implications more closely.

\section*{Acknowledgements}
This work is based on research sponsored by DARPA under agreement number FA8750-18-
2-0014. The U.S. Government is authorized to reproduce and distribute reprints for Governmental purposes notwithstanding any copyright notation thereon. The views and conclusions contained herein are those of the authors and should not be interpreted as necessarily representing the official policies or endorsements, either expressed or implied, of DARPA or the U.S. Government.

\bibliographystyle{coling}
\bibliography{coling2020}

\newpage
\setcounter{section}{0}
\renewcommand{\thesection}{\Alph{section}}
\section*{Appendix A: Additional Denoising Examples}

\begin{enumerate}
\item event, location of event
\begin{enumerate}
    \item Italy said on Monday it was sending 36 tonnes of medicines and other emergency aid to \textbf{Iran} after a massive \textbf{earthquake} struck part of the country's eastern region, killing about 2,400 people and injuring 6,000.
    \item Another \textbf{earthquake} hit \textbf{Iran} on Monday but was less severe than Saturday's quake, which measured 7.1 on the Richter scale.
    \item The United States announced on Monday that it would contribute \$100,000 to help victims of the \textbf{earthquake} in \textbf{Iran}.
\end{enumerate}
\item agent, event staged
\begin{enumerate}
    \item This fellow \textbf{Hun Sen}, chief perpetrator and one of communist Vietnam's puppets, staged a brutal military \textbf{coup}... after he returned from visiting his masters in Vietnam," the radio said.
    \item Cambodia's Second Prime Minister \textbf{Hun Sen} is effectively in control of the capital Phnom Penh and well entrenched in the south and east of the country after his virtual \textbf{coup} at the weekend, diplomats noted.
    \item It would be a de facto recognition of \textbf{Hun Sen}'s \textbf{coup} d'etat.
\end{enumerate}
\item designee, designation
\begin{enumerate}
    \item The mass of the people led by the liberation movement waged a just struggle against \textbf{apartheid}, which was designated by the United Nations as a \textbf{crime} against humanity.
    \item At the core of the ANC submission is the indictment of the system of \textbf{apartheid} as a \textbf{crime} against humanity...
\end{enumerate}
\end{enumerate}

\section*{Appendix B: Hyperparameter Search}

We experimented with BERT hidden layer dropout rates at fixed intervals of $10\%$ between $0\%$ and $50\%$.  We also experimented with every date window length between $0$ and $7$ when configuring our \textit{date-window} grouped and \textit{event-guided} entity pair selection, finding that a configuration which selected sentences up to 1) $4$ days after the event for all noun pairs and 2) $7$ days after the event for all named entity pairs (places, individuals, etc) led to the best performance.  When generating individual training batches, we used $6$ positive examples and $6$ negative examples, determined via grid search between $3$ and $10$.

\section*{Appendix C: Evaluation Corpora}
 
\begin{enumerate}
\item FewRel is a few-shot relation extraction corpus.  We evaluate on its development set to allow direct comparison with the results from~\newcite{Soares:19}.  The development set contains $16$ relations with $700$ examples per relation.  From this set, we generate $11,200$ test queries, using every example for every relation once.  The candidates are randomly sampled from the rest of the corpus.  To evaluate the performance of our Spanish training, we also translate 10\% of the examples for each relation in the FewRel development set with the Google Translate API.  While not professionally translated, it performs well on Spanish.  We thought the additional dataset was useful in evaluating the efficacy of our multilingual approach.  Reported numbers are the average of 10 trials.
\item SemEval 2010 Task 8 is a standard relation extraction corpus.  From its training set of $8,000$ examples, we fine-tune on just $80$ examples ($1\%$) with a learning rate of $3e-5$ and evaluate on both a held out development set of $1500$ examples ($\sim20\%$) and its test set of $2,717$ examples, performing similarly on both.  Our reported numbers are from the development set to allow direct comparison with~\newcite{Soares:19}.  Reported numbers are the average of 10 trials.
\item TAC KBP 2016 is a knowledge base population corpus for slot filling tasks in English and Spanish that contains source documents from discussion fora and news articles with accompanying gold standard relation labels (`physical', `orgaffiliation', `personalsocial', `partwhole', `generalaffiliation').  With a little bit of pre-processing to correct overlapping entities and filter unseparable entities (for example `los' and `los venezolanos'), we fine-tune on random fixed percentages ($1\%$, $10\%$, $20\%$) of its $2,403$ examples with a learning rate of $3e-5$ and evaluate on the remaining data.  Reported numbers are the average of 10 trials.
\end{enumerate}

\end{document}